\documentclass[conference]{IEEEtran}
\IEEEoverridecommandlockouts
\usepackage{cite}
\usepackage{amsmath,amssymb,amsfonts}
\usepackage{algorithmic}
\usepackage{graphicx}
\usepackage{textcomp}
\usepackage{xcolor}
\usepackage{subfig}
\usepackage{hyperref}
\def\BibTeX{{\rm B\kern-.05em{\sc i\kern-.025em b}\kern-.08em
    T\kern-.1667em\lower.7ex\hbox{E}\kern-.125emX}}

\usepackage{tikz}
\usepackage{textcomp}
\usepackage{hyperref}
\newcommand\copyrighttext{%
  \footnotesize \textcopyright~2025 IEEE. Personal use of this material is permitted. 
  Permission from IEEE must be obtained for all other uses, in any current or future 
  media, including reprinting/republishing this material for advertising or promotional 
  purposes, creating new collective works, for resale or redistribution to servers or lists, 
  or reuse of any copyrighted component of this work in other works. This is the author's version of the work. It has been accepted for presentation at the 
  2025 IEEE International Conference on Compute, Control, Network \& Photonics (ICCCNP). 
  The final version of record will be published in IEEE Xplore (DOI: to be added when available).%
}

\newcommand\copyrightnotice{%
  \begin{tikzpicture}[remember picture,overlay]
    \node[anchor=south,yshift=10pt] at (current page.south) {%
      \fbox{\parbox{\dimexpr\textwidth-\fboxsep-\fboxrule\relax}{\copyrighttext}}%
    };
  \end{tikzpicture}%
}

\begin{document}

\title{A Neural Network for the Identical Kuramoto Equation: Architectural Considerations and Performance Evaluation\\
}

\author{\IEEEauthorblockN{1\textsuperscript{st} Nishantak Panigrahi}
\IEEEauthorblockA{\textit{School of Computer Science and Engineering (SCOPE)} \\
\textit{Vellore Institute of Technology (VIT)}\\
Vellore, India \\
nishantak09@gmail.com,\\ nishantak.panigrahi2022@vitstudent.ac.in}
\and
\IEEEauthorblockN{2\textsuperscript{nd} Mayank Patwal}
\IEEEauthorblockA{\textit{School of Computer Science and Engineering (SCOPE)} \\
\textit{Vellore Institute of Technology (VIT)}\\
Vellore, India \\
mayanksingh4370@gmail.com,\\ mayanksingh.patwal2022@vitstudent.ac.in}
}

\maketitle
\copyrightnotice

\begin{abstract}
In this paper, we investigate the efficiency of Deep Neural Networks (DNNs) to approximate the solution of a nonlocal conservation law derived from the identical‐oscillator Kuramoto model, focusing on the evaluation of an  architectural choice and its impact on solution accuracy based on the energy norm and computation time. Through systematic experimentation, we demonstrate that network configuration parameters—specifically, activation function selection (tanh vs. sin vs. ReLU), network depth (4-8 hidden layers), width (64-256 neurons), and training methodology (collocation points, epoch count)—significantly influence convergence characteristics. We observe that tanh activation yields stable convergence across configurations, whereas sine activation can attain marginally lower errors and training times in isolated cases, but occasionally produce nonphysical artefacts. Our comparative analysis with traditional numerical methods shows that optimally configured DNNs offer competitive accuracy with notably different computational trade-offs. Furthermore, we identify fundamental limitations of standard feed-forward architectures when handling singular or piecewise-constant solutions, providing empirical evidence that such networks inherently oversmooth sharp features due to the natural function space limitations of standard activation functions. This work contributes to the growing body of research on neural network-based scientific computing by providing practitioners with empirical guidelines for DNN implementation while illuminating fundamental theoretical constraints that must be overcome to expand their applicability to more challenging physical systems with discontinuities.
\end{abstract}


\section{Introduction}

Partial Differential Equations (PDEs) are ubiquitous in modelling processes and their properties arising from a vast range of diverse fields, such as real-time engineering problems, different physical phenomena, chemical and biological events, and socio-economic theory~\cite{intro_bio}\cite{intro_socio}. Efficient and accurate solution of such complex dynamical systems becomes critical across a wide range of these applications. Classical numerical methods, such as finite-volume schemes can deliver accuracy but at the cost of long runtimes, especially for high-resolution simulations~\cite{intro_fvm}. Deep neural networks (DNNs) offer a mesh-free alternative by learning a continuous solution directly from the data known about the process, for example, the PDE. We enhance the DNN by embedding the PDE residual and boundary/initial-condition constraints into the network’s loss function, a paradigm henceforth referred to as a Physics-Informed Neural Network (PINN)~\cite{intro_pinn}. Consequently, we obtain a model that, once trained, produces solutions to the PDE almost instantaneously. {Furthermore, PINNs generalise better beyond the training domain and often exhibit faster training rates than traditional DNNs due to their incorporation of the governing physical laws directly into the loss function, thereby enforcing strong regularisation and consistency with known physics, and reducing reliance on large datasets~\cite{intro_ypinn}.  This makes PINNs particularly attractive for rapid computation tasks in mission-critical workflows. 

Despite these advances, the use of PINNs for nonlocal conservation laws that arise in a large variety of applications, particularly those derived from the Kuramoto model, remains underexplored. Moreover, deploying PINNs in real-world engineering requires a clear understanding of how architectural choices (e.g., activation functions, network depth/width) and training strategies (e.g., collocation sampling, epoch budgets) affect the solution fidelity.

In this paper, we fill this gap by conducting a comprehensive evaluation of DNN architectures by approximating the solution to the identical-oscillator Kuramoto Equation. Our contributions are threefold:

\begin{itemize}
    \item \textbf{Architectural analysis:} We compare $\tanh$, $\sin$, and $ReLU$ activations across network depths (4-8 layers) and widths (64-256 neurons), quantifying their influence on convergence and error. Highlighting cases where sine activations can yield marginal error reductions at the risk of nonphysical oscillations, and showcasing the failure of ReLU activation in capturing the evolution of the Kuramoto model.

    \item \textbf{Hyperparameter study:} We sweep collocation‐point sampling and epoch count, to identify configurations that optimise the trade-off between accuracy and training time.

    \item \textbf{Limitations and guidelines:} We demonstrate intrinsic oversmoothing of discontinuities by standard feed-forward DNNs and provide practical recommendations and empirical guidelines for practitioners.
\end{itemize}

\section{Physics-Informed Neural Network Formulation}
\subsection{Identical Kuramoto Equation}
We approximate the solution to the identical Kuramoto equation described in~\cite{kuram}, by approximating the phase density $u(\theta, t)$, satisfying the PDE
\begin{equation}
\frac{\partial u}{\partial t}(\theta,t)
+ \frac{\partial}{\partial \theta} \bigl[V[u](\theta,t)\,u(\theta,t)\bigr]
= 0, \;\; \theta \in [0, 2\pi], \ t \in [0, T]
\label{eq:pde}
\end{equation}
with smooth, polynomial initial condition described in \cite{kuram_fvm} as
$$
u(\theta,0) = u_0(\theta),\quad \theta\in[0,2\pi].
$$
\begin{equation}
u_0(\theta) = 
\begin{cases}
\displaystyle \frac{6}{\pi^3} \left( \frac{3\pi}{2} - \theta \right) \left( \theta - \frac{\pi}{2} \right), & \text{if } \theta \in \left[ \frac{\pi}{2}, \frac{3\pi}{2} \right], \\
0, & \text{otherwise}
\end{cases}
\label{eq:initial-condition}
\end{equation}

Where, the nonlocal flux is given by the convolution
\begin{equation}
V[u](\theta,t) = 
- K \int_{0}^{2\pi} \sin(\theta - \phi)\, u(\phi,t)\, \mathrm{d}\phi
\label{eq:kernel}
\end{equation}

For our simulation we have set the final time, $T=1$.

\subsection{Network Architecture}
A Deep Neural Network (DNN) is a parameterised function $u_\Phi:\mathbb{R}^d\to\mathbb{R}$, defined as a composition of affine maps and nonlinear activations. Effectively, our PINN \text{learns} the solution to the identical Kuramoto equation by a mapping from input $x{=}(\theta,t)~\in\mathbb{R}^2$, to an output $u(\theta, t)\in\mathbb{R}$.

The PINN, $\mathcal{N}_\Phi$, is constructed as a fully-connected feed-forward network, as shown in Fig.~\ref{net}, with an input layer of dimension 2 (receiving spatial and temporal coordinates), $L$ hidden layers, each with $n$ neurons, with nonlinear activation functions $\sigma$, and a linear output layer that yields the approximation $u(\theta, t)$,
\begin{figure}[htbp]
\centering{\includegraphics[scale=0.3]{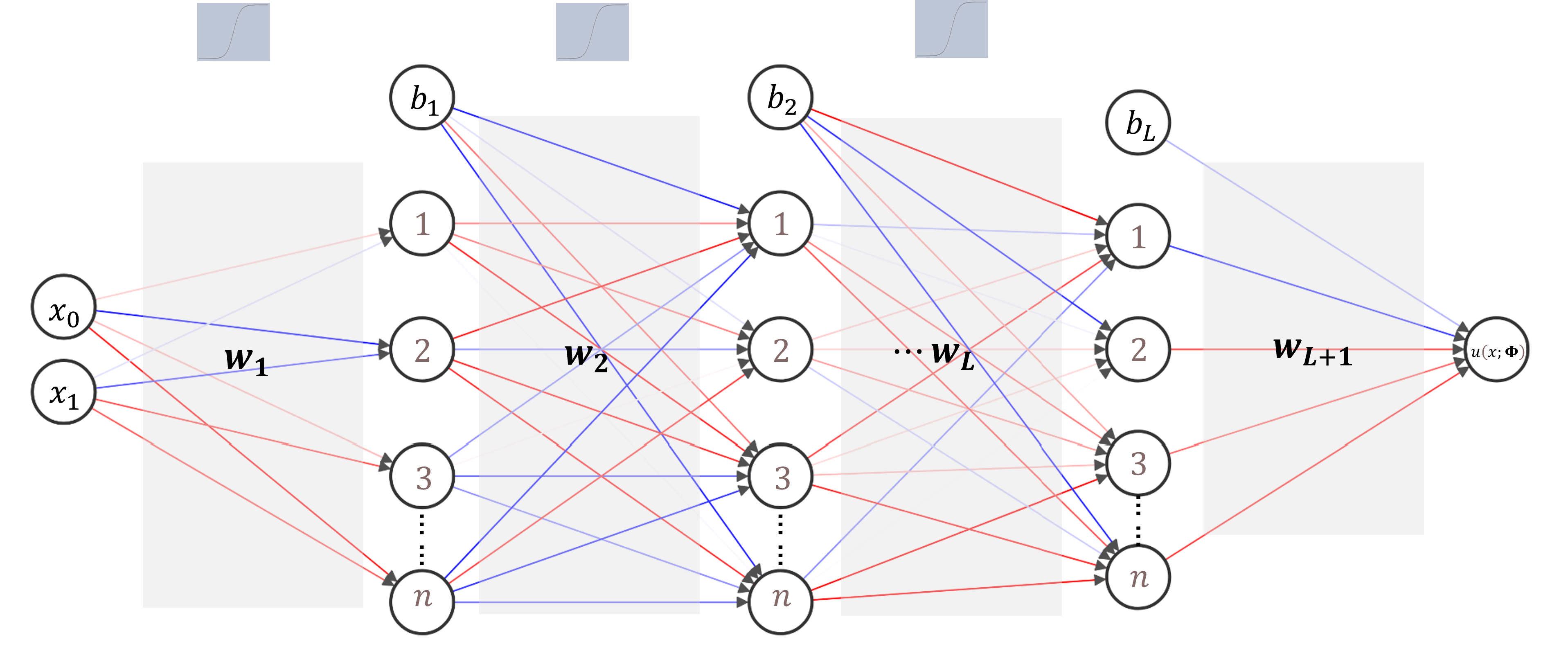}}
\caption{Deep neural network architecture.}
\label{net}
\end{figure}

It is formulated as
$$
\mathcal{N}(x; \Phi): \mathbb{R}^2 \to \mathbb{R}\\
$$
For our architecture we have chosen $L{=}4,\; n{=}64, and $ 
$\sigma{=}\tanh$. This gives us
$$
\begin{aligned}
x &= [\,\theta,\,t\,]^\top,\\
\mathbf{h}^{(0)} &= x,\\
\mathbf{h}^{(i)} &= \tanh\bigl(W^{(i)}\,\mathbf{h}^{(i-1)} + \mathbf{b}^{(i)}\bigr),\quad i=1,\dots,4,\\
\mathcal{N}(x; \Phi) &= W^{5}\,\mathbf{h}^{(4)} + \mathbf{b}^{(5)} = u(\theta, t)
\end{aligned}
$$ 
The trainable parameters, $\Phi=\{W^{(i)},\;\mathbf{b}^{(i)}\}_{i=1}^{5}$, are learnt by a minimising a scalar loss $L(\Phi)$.\\
Where for layer $i$ with $n_i$ neurons,
\begin{itemize}
    \item $W^{(i)}\in\mathbb{R}^{n_i\times n_{i-1}}$ is the weight matrix,

    \item $\mathbf{b}^{(i)}\in\mathbb{R}^{n_i}$ is bias matrix,
\end{itemize}

\subsection{Loss Components}
\subsubsection{PDE Residual}

Defines the point-wise residual of our PDE
\begin{equation}
r(\theta,t;\ \Phi) = 
\frac{\partial u_\Phi}{\partial t} 
+ \frac{\partial}{\partial \theta} \bigl( V[u_\Phi]\, u_\Phi \bigr)
\label{eq:residual}
\end{equation}

Sampling $N_r{=}1024$ collocation points $(\theta_i, t_i)$, drawn uniformly via Latin‑Hypercube sampling over $[0, 2\pi]\times[0, T]$, we form
\begin{equation}
L_{\mathrm{res}} = 
\frac{1}{N_r} \sum_{i=1}^{N_r} r(\theta_i, t_i;\ \Phi)^2
\label{eq:loss}
\end{equation}

Nonlocal Flux: Nonlocal convolution, $V[u]$ is computed via a Riemann sum. We introduce $N_q{=}128$ uniformly spaced nodes $\{\phi_j\}_{j=1}^{N_q}$ on $[0,2\pi]$ with step size $\Delta\phi = 2\pi/N_q$. The integral is then approximated as
\begin{equation}
V[u_\Phi](\theta_i, t_i) \approx 
- K \sum_{j=1}^{N_q} \sin(\theta_i - \phi_j)\, u_\Phi(\phi_j, t_i)\, \Delta\phi
\label{eq:velocity-discrete}
\end{equation}

\subsubsection{Initial Condition Loss}
We enforce $u_\Phi(\theta, 0) = u_0(\theta)$ by sampling $N_0{=}512$ initial condition points, $\{\theta_j^0\}_{j=1}^{N_0}$ over $[0, 2\pi]$, and computing
\begin{equation}
L_{\mathrm{IC}} = 
\frac{1}{N_0} \sum_{j=1}^{N_0} \left| u_\Phi(\theta_j^0, 0) - u_0(\theta_j^0) \right|^2
\label{eq:loss-ic}
\end{equation}

\subsubsection{Total Loss}
The total physics-informed loss that the network sees becomes a weighted sum
$$
L_{\mathrm{total}}
\;=\;\lambda_{res}\cdot L_{\mathrm{res}} \;+\; \lambda_{IC}\cdot L_{\mathrm{IC}}.
$$
where $\lambda_{res},\;\lambda_{IC}$ are the weights of the residual loss and the initial condition loss respectively. In our implementation we have set $\lambda_{res}=\lambda_{IC}=1$.

\subsection{Training and Optimisation}

\begin{itemize}
    \item \textbf{Automatic Differentiation:} All derivatives $\partial_tu_\Phi$ and $\partial_\theta(\mathbf{F}_\Phi)$ are computed via the framework’s auto‐grad engine, which is significantly faster than numerical differentiation and {one of the motivations behind using neural networks for scientific computing~\cite{motiv_pinn}.}

    \item \textbf{Optimiser:} Model Pprameters are updated using backpropagation gradients by the Adam optimiser, with learning rate $10^{-3}$, typically over $4096$ or $5120$ epochs.
\end{itemize}

\section{Experimental Setup \& Methodology}
In the following experiments, simulation parameters are purely synthetic, approximating different order of smoothnesses occurring in nature.

\subsection{Network Architecture and Hyperparameter Grid}
We train and evaluate fully-connected, feed‑forward DNNs by systematically varying the architecture as follows:

\begin{itemize}
    \item Activation function, $\sigma\in\{\tanh, sine, ReLU\}$ 
    \item Network depth, $L$, and width, $n$,
    $$
    \{L, n\}\in\{\{4,64\}, \{4,128\}, \{6,128\}, \{6, 256\}, \{8,256\}\}
    $$
\end{itemize}
We vary training hyperparameters as follows:
\begin{itemize}
    \item Epoch counts, $n_{e}\in\{2048, 4096, 5120, 10240\}$
    \item Collocation points, $N_r\in\{1024, 2048\}$
\end{itemize}
Each trained network, $\mathcal{N}_\Phi^{\{\sigma, \{L, n\}, n_{e}, N_r\}}$, learns the trainable parameters $\Phi$ to map an input $x=\{\theta, t\}$ to the output $u_\Phi(\theta, t)$, passing it through $L$ hidden layers of size $n$, and a linear output layer.

\subsection{Evaluation Metrics}
\subsubsection{Exact Solution}
We compute a high-fidelity solution, $u_{ex}(\theta, t)$, using a high-resolution finite-volume solver described in \cite{kuram_fvm}. 

A super-fine mesh solution with $512$ spatial points, and $205$ temporal points over the domain, is saved to serve as the reference solution for evaluating and comparing the accuracy of the PINN solutions. The method uses a finite-volume scheme with Lax-Friedrichs numerical flux.

\subsubsection{Energy Norm}
We quantify the solution accuracy by calculating the discrete $L^2$ norm
\begin{equation}
\| u_\Phi - u_{\mathrm{ex}} \|_E = 
\sqrt{ \frac{1}{N_{\mathrm{eval}}} 
\sum_{k=1}^{N_{\mathrm{eval}}} 
\left( u_\Phi(\theta_k, t_k) - u_{\mathrm{ex}}(\theta_k, t_k) \right)^2 }
\label{eq:error-eval}
\end{equation}

This gives us a single scalar measuring the deviation between the network’s prediction and the high‐fidelity reference solution. Physically,$ \|u_\Phi - u_{ex}\|_E$ represents the \textit{total energy} of the error field, accumulated over all phases and times, giving us a metric for direct comparison of architectures.

\subsubsection{Training Time}
The total wall-clock time each network configuration takes to finish training is measured in \textit{seconds} using Python's \textit{time} module, and serves as a direct metric of computational efficiency across the grid.

\subsection{Training Hardware}
All experiments are conducted on a laptop configured with an AMD Ryzen 7 7735HS CPU, 16 GB of system memory, and an NVIDIA RTX 3050 dGPU (6 GB VRAM), running CUDA 12.8. Neural network training is performed on the dGPU via PyTorch’s CUDA backend, with a fixed seed to ensure consistency and reproducibility of timing measurements.

\section{Results and Discussions}
In this section, we present a quantitative evaluation of the trained networks across our architectural and hyperparameter grid, and analyse the effect of varying activation function, network depth and width, collocation density, and epoch budget, on solution accuracy and training efficiency. 

\subsection{Activation Functions}
Networks with hyperbolic‐tangent activation, $\tanh$, consistently demonstrate inherently reliable convergence in smooth function appropriators. The best-performing configuration $\sigma=\tanh, \{L{=}4, n{=}128\}, N_r{=}1024, n_e{=}4096$ showed the least energy norm, $RMSE = 6.65\times10^{-5}$ with a training time of $\sim114~seconds$. The evolution of the solution is shown in Fig.~\ref{tan_soln}.
Approximately $5\%$ of $\tanh$ networks showed nonphysical artefacts. Importantly, these occurrences were confined to configurations with excessive depth and width, suggesting that over‐parameterisation for a simple PDE leads the model to fit spurious features rather than generalise the true solution structure. 
\begin{figure}[htbp]
\centering{\includegraphics[scale=0.22]{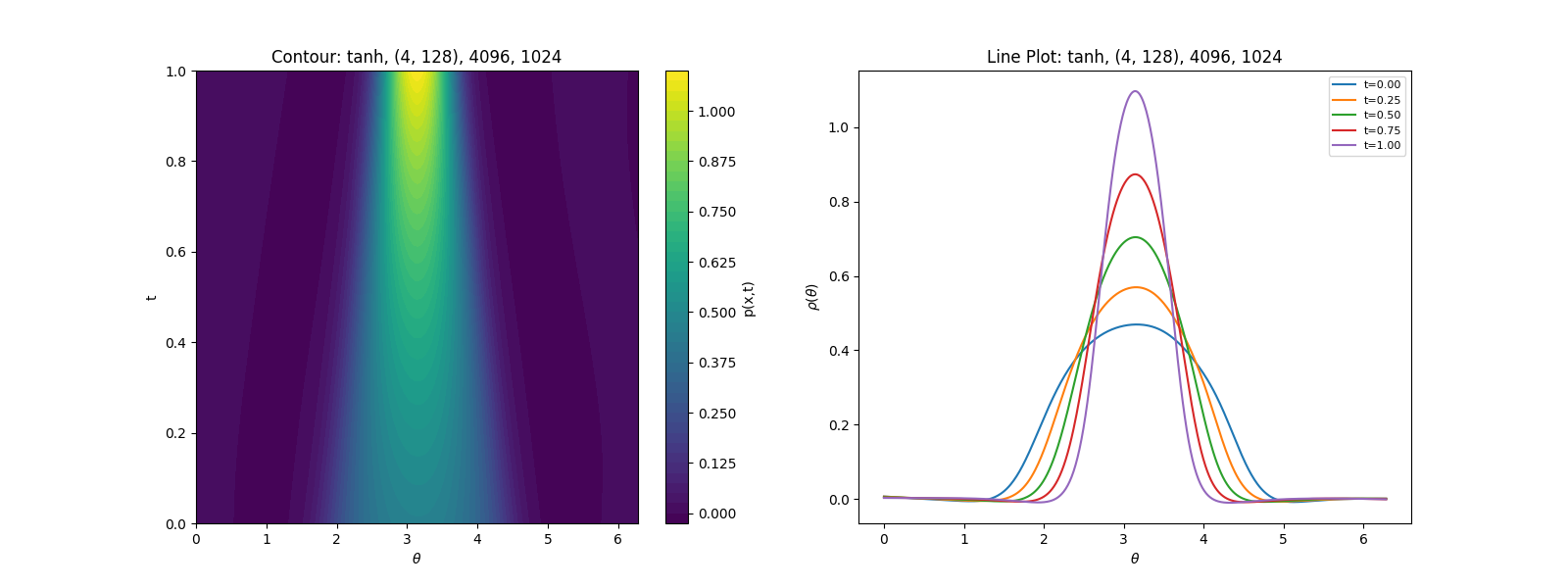}}
\caption{Simulation profile of $\tanh$ network.}
\label{tan_soln}
\end{figure}

Best-case trials for $\sin$ networks occasionally yielded marginally lower energy norm for the same architecture with $\tanh$ activation, showing upto $10\%$ lower energy norms, and approximately $13\%$ faster convergence than $\tanh$ architectures with similar energy norm, as shown in Fig.~\ref{tt_l2}. The evolution of the solution is shown in Fig.~\ref{sin_soln}. However, such networks produced nonphysical artefacts, as shown in Fig.~\ref{nonphy}, in approximately $10\%$ of configurations. These artefacts were, predominantly, spurious high‑frequency oscillations which can be attributed to the intrinsic oscillatory nature of the sine activation, whose nondecaying, sign‑alternating derivative, cosine, amplifies unresolved spectral modes under insufficient collocation density, a behaviour not observed with $\tanh$, whose monotonic, low‑pass characteristic and bounded derivative dampen such perturbations.
\begin{figure}[htbp]
\centering{\includegraphics[scale=0.35]{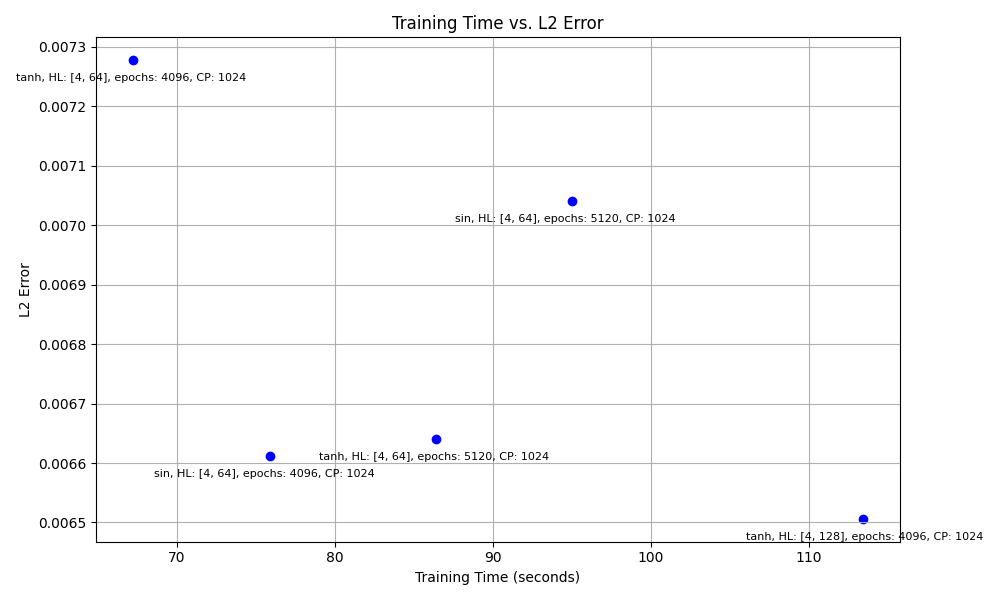}}
\caption{Training time vs. $L^2$ error of $\sin$ and $\tanh$ networks}
\label{tt_l2}
\end{figure}

\begin{figure}[htbp]
\centering{\includegraphics[scale=0.22]{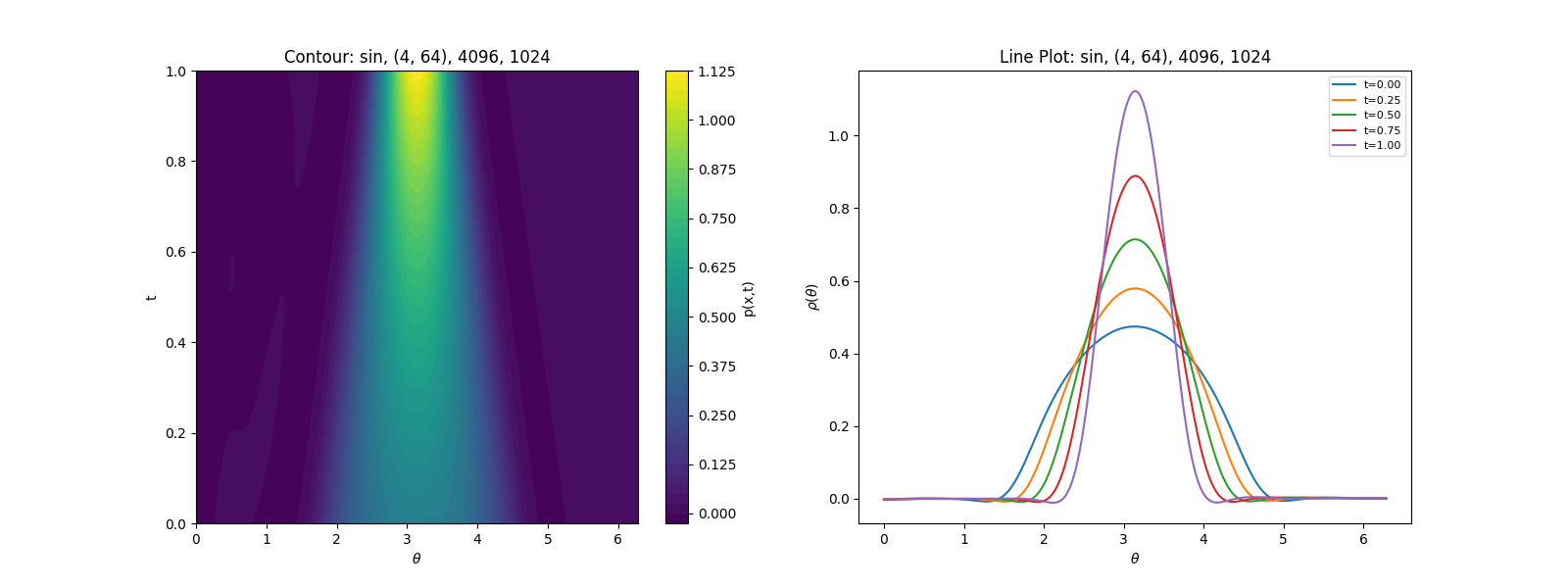}}
\caption{Simulation profile of $\sin$ network.}
\label{sin_soln}
\end{figure}

\begin{figure}[htbp]
\centering{\includegraphics[scale=0.22]{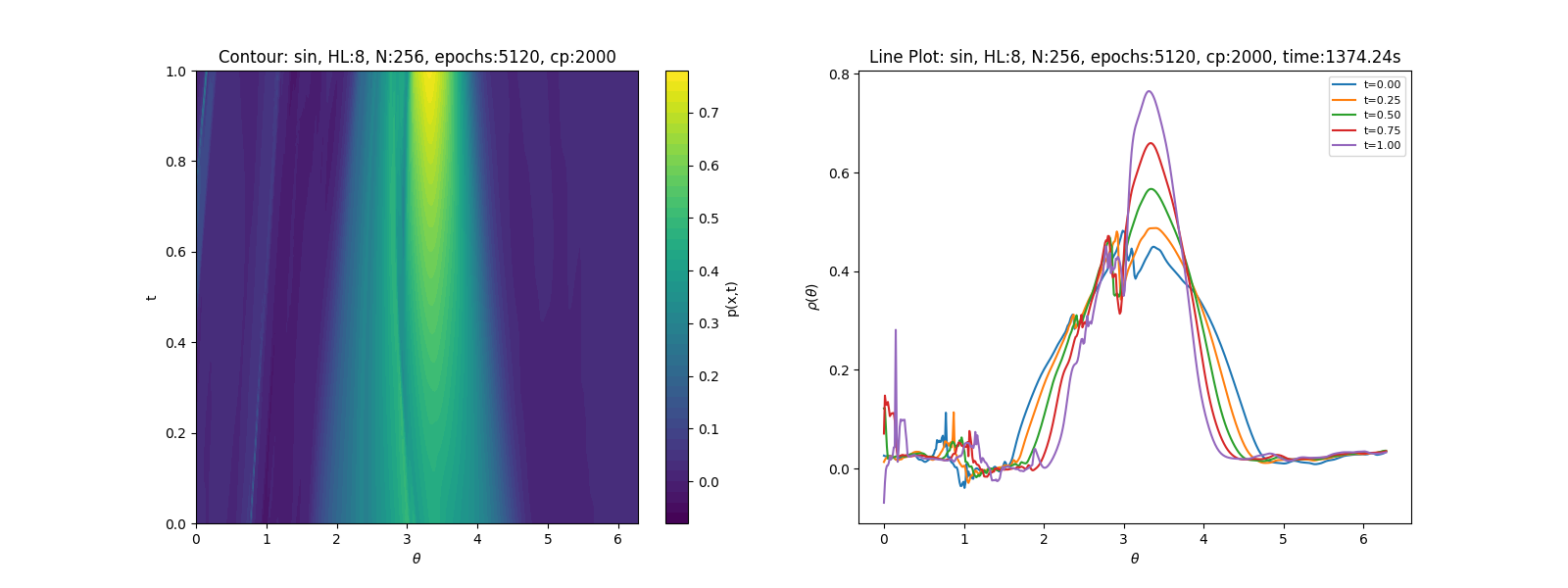}}
\caption{Nonphysical artefacts in a $\sin$ network simulation profile.}
\label{nonphy}
\end{figure}

In contrast, Fig.~\ref{relu} shows that $ReLU$ networks consistently fail to capture the solution across all architectures and hyperparameters. These results confirm that smooth activation functions, such as $\tanh$, $\sin$ are essential to scientific computing using DNNs. 
\begin{figure}[htbp]
\centering{\includegraphics[scale=0.35]{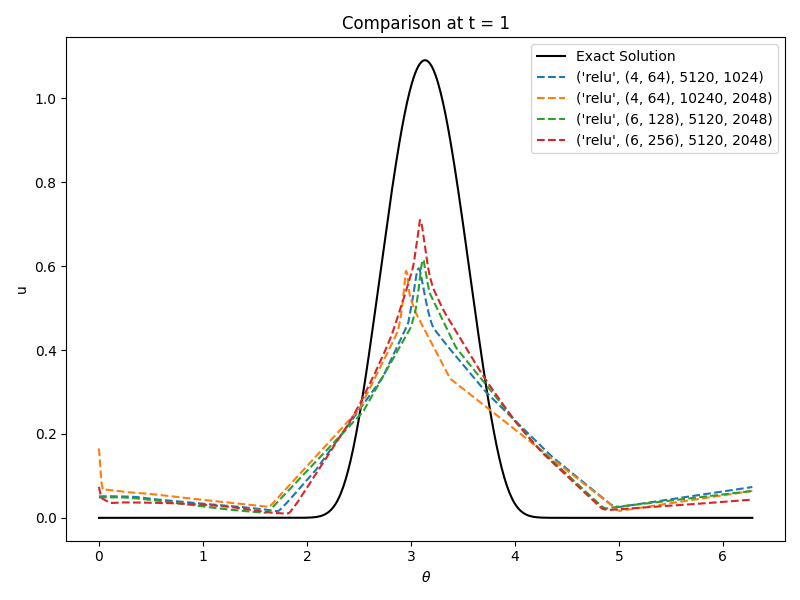}}
\caption{Solutions of ReLU networks.}
\label{relu}
\end{figure}

\subsection{Network Depth and Width}
We vary network depth and width in a $\tanh$ network and quantify accuracy and efficiency, as shown in Fig.~\ref{comp_tan}.

\begin{figure}[htbp]
\centering{\includegraphics[scale=0.29]{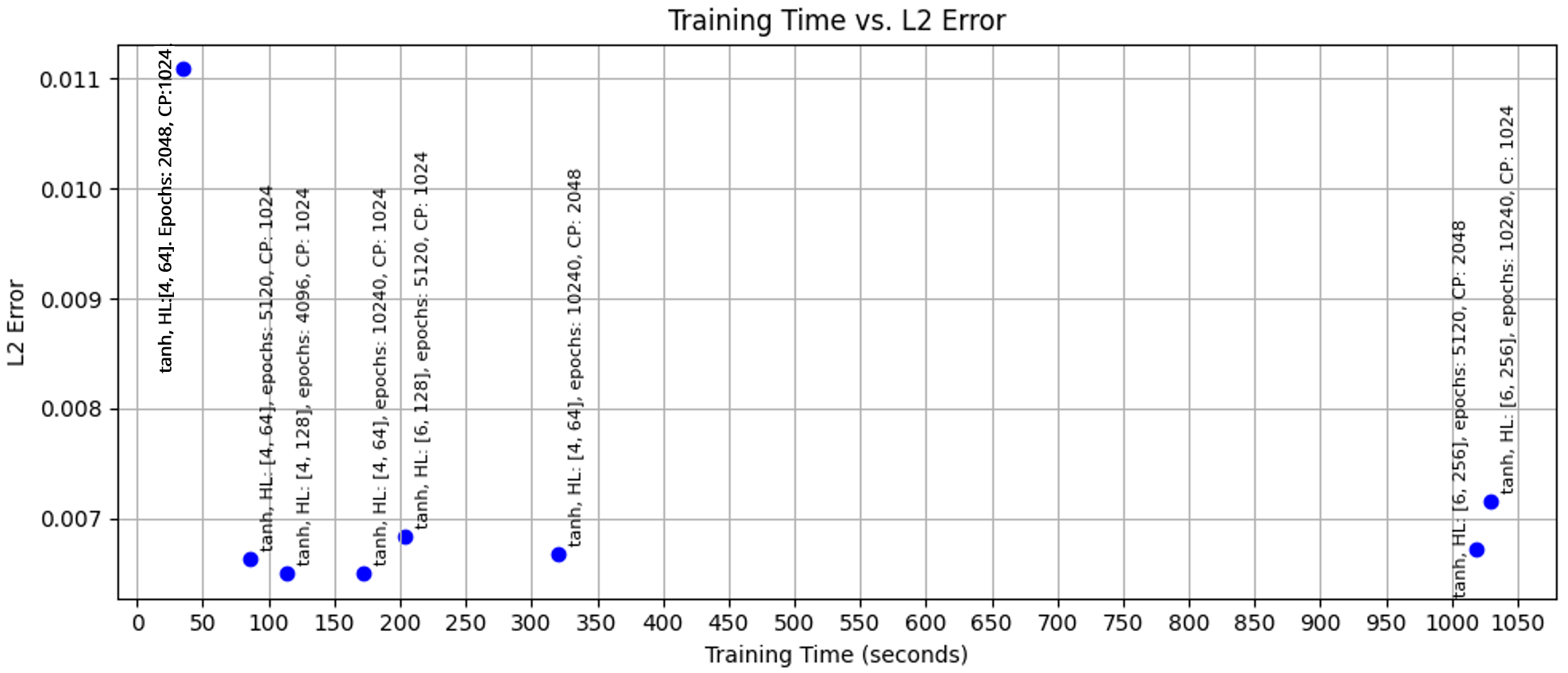}}
\caption{Training time vs. $L^2$ error of $\tanh$ networks.}
\label{comp_tan}
\end{figure}

The configuration $\{\{L{=}4, n{=}64\}, N_r{=}1024, n_e{=}2048\}$ completes training in only $35~s$ but incurs a relatively high error of $1.11\times10^{-4}$. Doubling the epoch budget to $n_e=4096$ for the same architecture yields approximately a twofold reduction in error ($\sim7\times10^{-5}$) for a modest increase in training time ($\sim70~s$).

Increasing the width from $n{=}64$ to $n{=}128$ (with $L{=}4$, $N_r{=}1024$, $n_e{=}4096$) raises the training time to $\sim114~s$ but further reduces error down to $6.65\times10^{-5}$. Demonstrating that in simple PDEs, width scaling may be more effective in capturing the solution dynamics than simply extending epochs.

Epoch budgets, $n_e$, beyond $5120$ produces diminishing returns but show significant increases in training times, indicating that the networks converged in $n_e\le5120$. To avoid wasted computations, it is essential to monitor convergence of the training loss during epochs along with techniques like early-stopping.

Deepening to $L{=}6$ and widening to $n{=}256$ dramatically increases training time to $>1000 s$, while offering negligible error reduction. This confirms that large feed-forward architectures do not offer any benefits over smaller networks, for simple PDEs. However, for simple PDEs caution must be exercised as over-parameterisation may lead to \textit{memorisation} rather than generalisation.

Overall, the Pareto‑front of minimal training time versus minimal $L^2$ error is traced by shallow ($L{=}4$), moderately wide ($n = 64{-}128$) networks with epoch counts, $n_e$, in the range $4096{-}5120$, giving the best balance between convergence speed and accuracy. The solutions produced by these architectures are shown in Fig.~\ref{all}, in contrast to a ReLU-based plot.

\begin{figure}[htbp]
\centering{\includegraphics[scale=0.35]{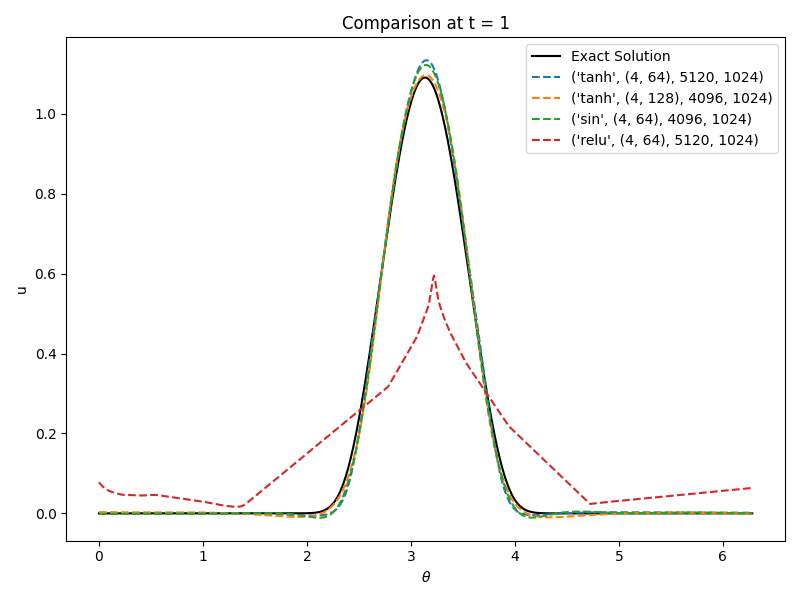}}
\caption{Solutions of trained DNNs against reference solution.}
\label{all}
\end{figure}

\subsection{Collocation Density}
Fig.~\ref{comp_tan} shows that raising the collocation density from $N_r{=}1024$ to $N_r{=}2048$ does not produce any substantial accuracy gains unless accompanied by either prolonged training or a larger network. Even then the gains are marginal (${<}10\%$). This behaviour suggests that denser sampling alone cannot improve the solution unless the network has sufficient representational capacity and training time to exploit the extra information.  Empirically, $N_r{=}1024$ consistently struck a good balance between most of the accuracy benefits while keeping the wall‑clock cost low in typical GPU-laptop runtimes, particularly for simple PDEs.

\subsection{Limitations: Discontinuous Initial Data}
While standard feed‑forward PINNs demonstrate competitive accuracy on smooth initial conditions, our experiments reveal fundamental shortcomings when the initial data contains singularities or piecewise-constant functions. We explore these shortcomings by training several architectures and activation functions on 

1. Dirac-delta initial condition described in \cite{kuram_fvm} as
\begin{equation}
\delta_\varepsilon(x - a) =
\begin{cases}
\dfrac{1}{2\varepsilon}, & \text{if } |x - a| < \varepsilon, \\[7pt]
0, & \text{otherwise},
\end{cases}
\qquad
H(x) =
\begin{cases}
1, & \text{if } x \ge 0, \\
0, & \text{if } x < 0.
\end{cases}
\label{eq:delta-heaviside}
\end{equation}

\begin{equation}
\begin{split}
u_0(\theta) &= \frac{1}{4} \bigl[\delta_\varepsilon\bigl(\theta - \tfrac{3\pi}{4}\bigr)
+ \delta_\varepsilon\bigl(\theta - \tfrac{5\pi}{4}\bigr)\bigr] \\
&\quad + \frac{1}{2} H\bigl(\theta - \tfrac{\pi}{2}\bigr)
\bigl[1 - H\bigl(\theta - \tfrac{3\pi}{2}\bigr)\bigr]
\end{split}
\end{equation}

The closest solutions evolved as shown in Fig.~\ref{singular}. \\
\begin{figure}[htbp]
    \centering
    \subfloat[$\sin$ activation]{%
        \includegraphics[scale=0.22]{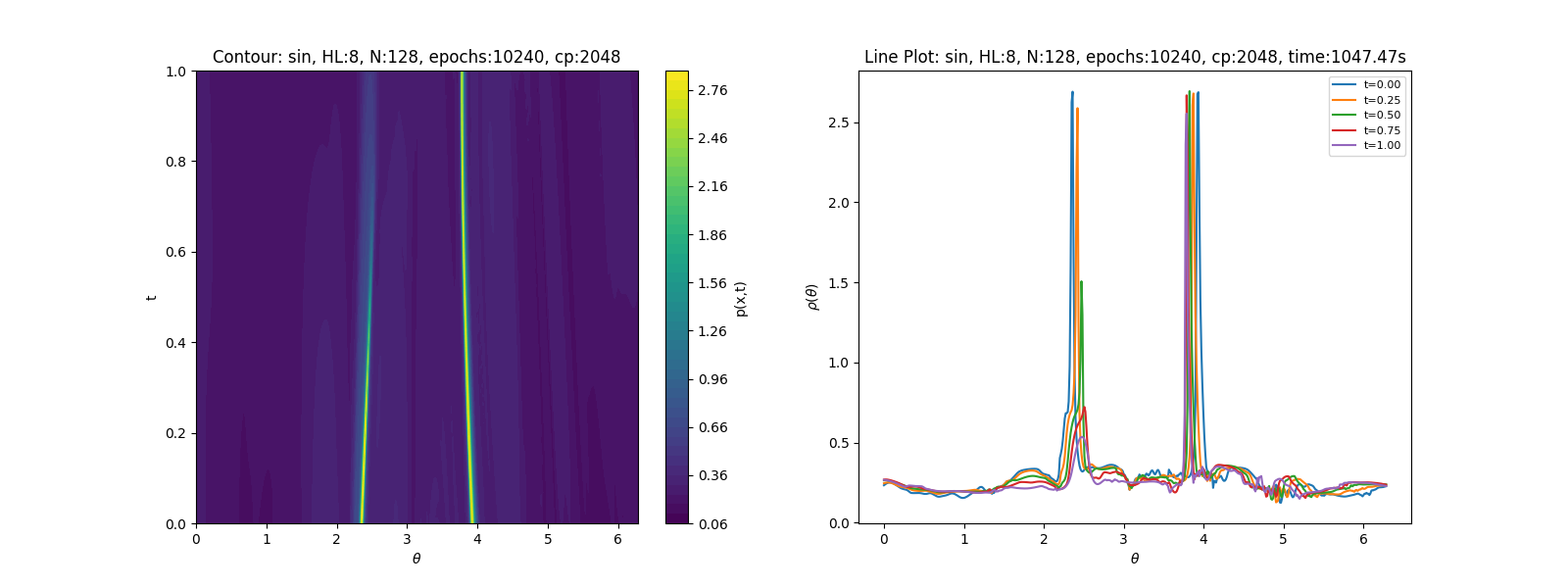}
        \label{fig:sin}
    }
    \hfil
    \subfloat[$\tanh$ activation]{%
        \includegraphics[scale=0.22]{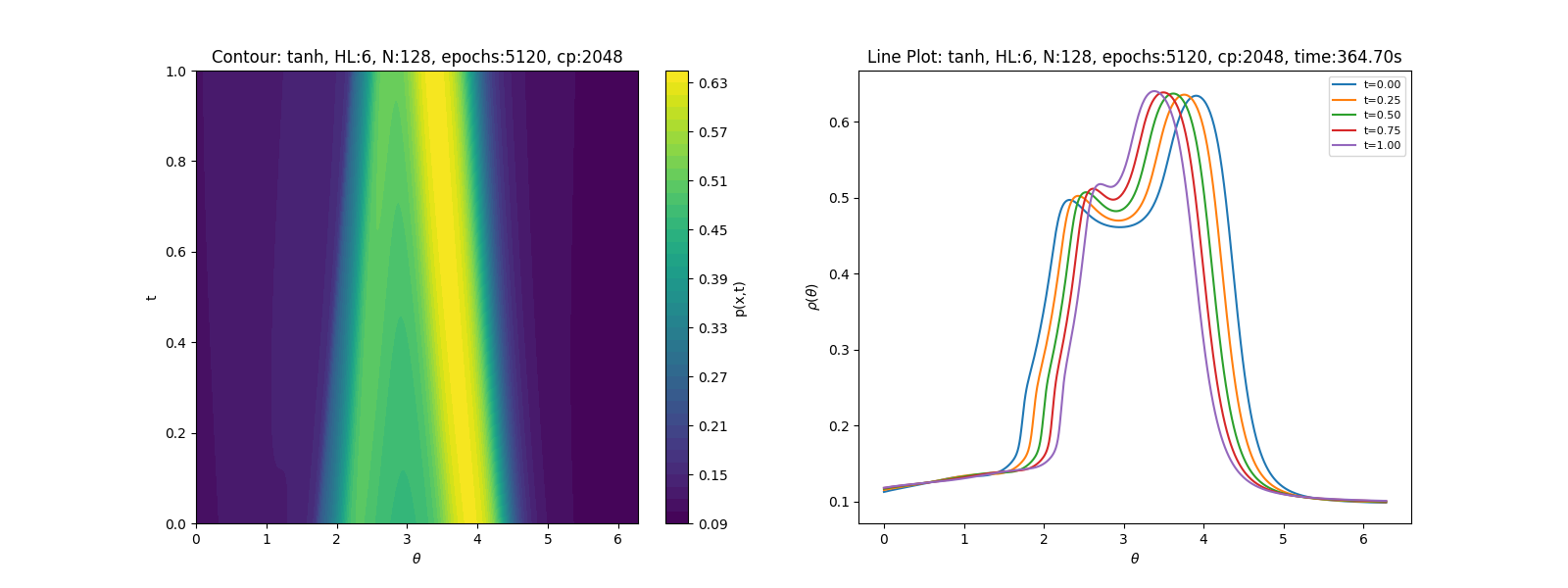}
        \label{fig:tanh}
    }
    \caption{Simulation profiles of the closest solutions for initial data with singular masses.}
    \label{singular}
\end{figure}

2. Piecewise-constant initial condition described in \cite{kuram_fvm} as
\begin{equation}
u_0(\theta) =
\begin{cases}
\dfrac{2}{3\pi}, & \text{if } \frac{\pi}{2} \le \theta \le \frac{3\pi}{2}, \\[8pt]
\dfrac{1}{3\pi}, & \text{otherwise}.
\end{cases}
\label{eq:u0-piecewise}
\end{equation}

The closest solutions evolved as shown in Fig.~\ref{piecewise}. \\
\begin{figure}[htbp]
    \centering
    \subfloat[$\sin$ activation]{%
        \includegraphics[scale=0.22]{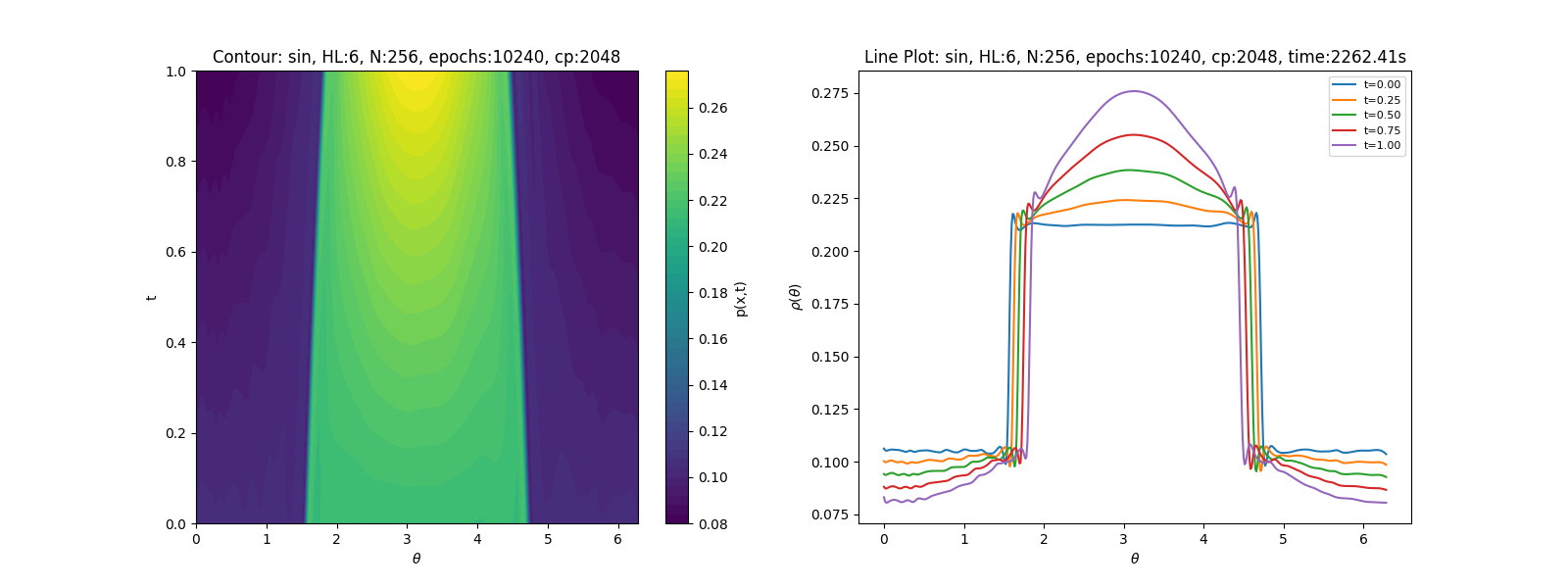}
        \label{fig:sin_piecewise}
    }
    \hfil
    \subfloat[ReLU activation]{%
        \includegraphics[scale=0.22]{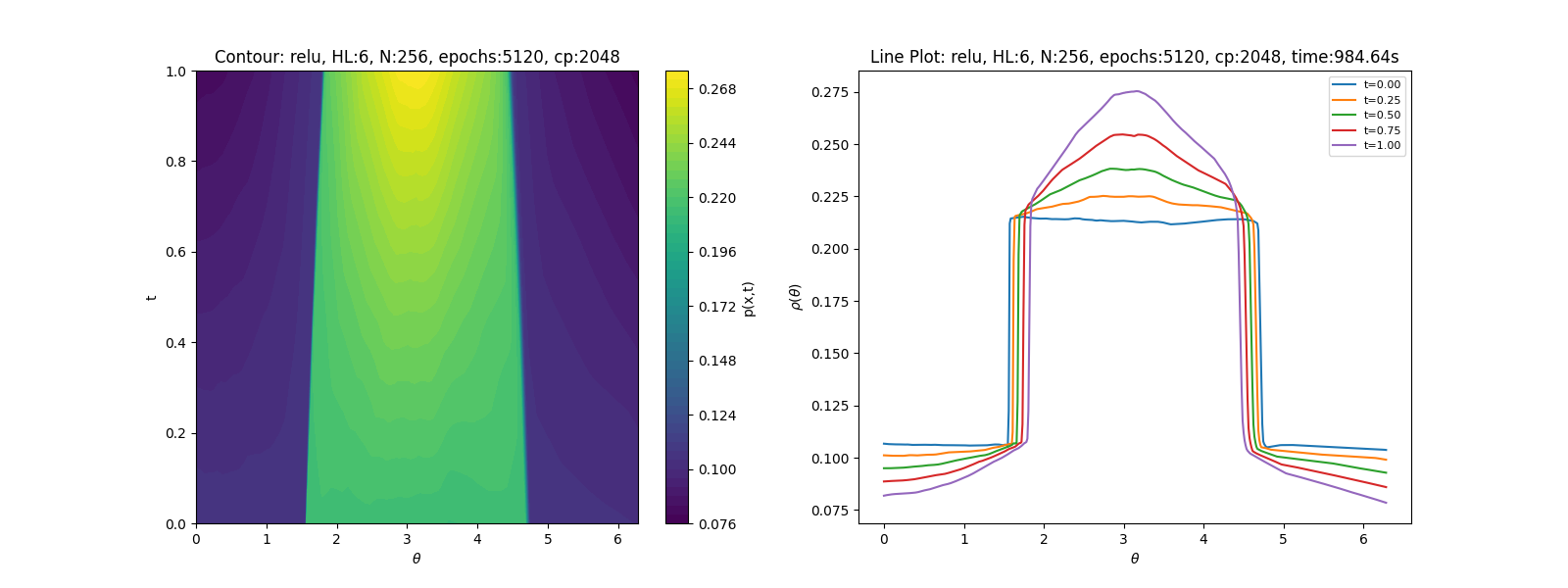}
        \label{fig:relu_piecewise}
    }
    \caption{Simulation profiles of the closest solutions for initial data with discontinuities.}
    \label{piecewise}
\end{figure}

We find that all tested configurations oversmooth the sharp jumps and discontinuities present in these initial conditions. The output approximate solution looks like a smooth function, as shown in Fig.~\ref{singular}(b), dampening these spikes. The piecewise-constant data is approximated more accurately than the singular data, by networks using $\sin$ and $ReLU$ activations, due to the oscillatory behaviour of $\sin$ and the piecewise-linear nature of ReLU. However, as shown in Fig.~\ref{piecewise}, both networks exhibit spurious oscillations near discontinuities—a Gibbs phenomenon-like artefact, arising from the difficulty of representing abrupt changes by the network.

However, this is consistent with neural-network behaviour because of:
\begin{itemize}
    \item \textbf{Smooth Activation Bias:} As discussed in Section 6.2 of~\cite{dl_book}, and theoretically analysed in~\cite{fail_ntk} smooth activation functions used in DNNs for scientific computing (e.g., $\tanh,\; \sin$) enforce $C^\infty$ solution spaces. This means that the resulting composite map
    \begin{equation}
    u_\Phi(\theta, t) =
    W^{(L+1)}\, \sigma\bigl( W^{(L)} \bigl( \cdots 
    \sigma( W^{(1)} [\theta, t] ) \cdots \bigr) \bigr)
    \label{eq:nn-output}
    \end{equation}
    
    is in $C^\infty$ class, i.e., an infinitely differentiable function. Therefore, it cannot have a true discontinuity or delta-like spikes. Any such feature would require an unbounded derivative, which at that point would blow up.

    \item \textbf{Strong formulation of PDE:} A standard PINN sees the PDE in its strong (point-wise) form, $r(\theta_i,t_j)$, and then trains to minimise $MSE(r(\theta_i, t_j)) \to 0$. Automatic differentiation using auto-grad gives us the strong derivative of a smooth network even when the true solution has weak derivative or a distributional derivative like a delta function. At the location of a jump, $u_\Phi$ smoothly interpolates the discontinuity, thereby misinterpreting the physics of the discontinuity~\cite{disco}. Hence, there is no large, localised penalty at that interface and the optimiser satisfies $r{\approx}0$ in the $MSE$ most easily by converging to a lower-curvature surrogate that smoothens the discontinuities.
\end{itemize}

\subsection{Remedies for Discontinuities and Spikes}
To mitigate the inherent oversmoothing of discontinuities by standard feed‑forward DNNs, future studies and applications should look into any of the following techniques: weak or variational loss formulations that integrate the residual against test functions~\cite{wpinn}, thereby providing distributional derivatives and gradients at jumps. Domain decomposition by training separate subnetworks on subintervals~\cite{dom} so as to localise shocks and prevent global smoothening; this would however require a priori knowledge of domain and solution structure. Enriched input embeddings, e.g., Fourier features~\cite{fourier} or wavelets may further enhance the network’s expressivity to capture steep gradients. One emerging field is quantum machine learning, wherein Parameterised Quantum Circuits (PQCs) use quantum effects to capture nonlocal correlations~\cite{quantum}. Work in these techniques can extend DNNs’ applicability to problems featuring discontinuities and singular measures.

\section{Conclusion}
Within this work, we have presented a systematic and comprehensive study of feed-forward DNN architectures, training styles, and their inherent limitations. By applying a PINN to a nonlocal conservation law - the identical-oscillator Kuramoto equation, and performing an extensive grid search over activation functions, network depth and width, collocation density, and epoch budgets, we identified that moderately-sized networks with four to six hidden layers of $64$ to $128$ neurons each, with $\tanh$ activation, prove to be the most stable and provide the most competitive accuracy, and training times.

We have empirically demonstrated the limitations of standard DNNs in approximating functions with discontinuities and singular masses due the combined effect of smooth activation function and auto-grad on strong-form PDE residual. To overcome these shortcomings we have outlined possible future directions of research with promising scope. 

Our experiments and findings, thus clearly lay empirical foundations for architectural considerations in DNN models for scientific computing. Practitioners must heed these guidelines and, where necessary, adopt specialised remedies to overcome the limitations identified. Therefore, this work effectively enriches the current body of literature on the use of DNNs for scientific and industrial applications.

\end{document}